\documentclass[journal,transmag]{IEEEtran}
\ifCLASSOPTIONcompsoc
  \usepackage[nocompress]{cite}
\else
  \usepackage{cite}
\fi

\ifCLASSINFOpdf

\usepackage{amsmath}
\usepackage{graphicx}
\usepackage{esint}
\usepackage{array}
\usepackage{multirow}
\usepackage{color}
\usepackage{epstopdf}
\usepackage{textcomp}
\begin{document}
\title{Improved Preterm Prediction Based on Optimized Synthetic Sampling of EHG Signal }

\author{Jinshan~Xu,
	Zhenqin~Chen,
	Yanpei~Lu,
	Xi~Yang,
	Alain~Pumir
	\IEEEcompsocitemizethanks{\IEEEcompsocthanksitem   Manuscript received xxxx.
This work is supported by the National Natural Science Foundation of China under grants 61873238, U1509207, 6160339, 61672463,
and Zhejiang Provincial Natural Science Foundation of China under grant LY20F030018.
	
J. Xu,  Z. Chen, Y. Lu, \& X. Yang are with College of Computer Science and Technology,
Zhejiang University of Technology, Hangzhou, 310023, China.
A. Pumir is with the Laboratoire de Physique, Ecole Normale Sup\'{e}rieure de Lyon, Lyon, 69007, France.
Corresponding author: X. Yang  (email: xyang@zjut.edu.cn)
	}% <-this % stops an unwanted space
	%\thanks{Manuscript received \today; revised  \today.}
	}

\markboth{}%
{Shell \MakeLowercase{\textit{et al.}}: Bare Demo of IEEEtran.cls for IEEE Transactions on Magnetics Journals}

\IEEEtitleabstractindextext{%
\begin{abstract}
Preterm labor is the leading cause of neonatal morbidity and mortality and has attracted research efforts from many scientific areas. The inter-relationship between uterine contraction and the underlying electrical activities makes uterine electrohysterogram (EHG) a promising direction for preterm detection and prediction. Due the scarcity of EHG signals, especially those of preterm patients, synthetic algorithms are applied to create artificial samples of preterm type in order to remove prediction bias towards {term}, at  the expense of a reduction of the feature effectiveness in machine-learning based automatic preterm detecting. To address such problem, we quantify the effect of synthetic samples (balance coefficient) on features' effectiveness, and form a general performance metric by utilizing multiple feature scores with relevant weights that describe their contributions to class separation.  Combined with the activation/inactivation functions that characterizes the effect of the abundance of training samples in term and preterm prediction precision, we  obtain an optimal sample balance coefficient that  compromise the effect of synthetic samples in removing bias towards the {majority} and the side-effect of  reducing features' importance.  Substantial improvement  in prediction precision  has been achieved through a set of numerical tests on public available  TPEHG database,   and it verifies the effectiveness of the proposed method.
\end{abstract}

\begin{IEEEkeywords}
preterm prediction, uterine electrohysterogram, synthetic sampling, sample balance coefficient
\end{IEEEkeywords}}

\maketitle

\IEEEdisplaynontitleabstractindextext

\IEEEpeerreviewmaketitle

\section{Introduction} \label{sec:introduction}

\IEEEPARstart{P}{reterm} birth, defined as babies born with gestation age less than 37 weeks,
is a major and growing challenge for public health systems.
Nearly 15 millions babies, or about 10\% of total world-wide births, are born prematurely each year.
About one million of these preterm babies die because of complications following the preterm birth~\cite{Howson2013}.
%The current lack of a comprehensive understanding of the mechanism of the initiation of uterine contraction  hinders any effective treatment of preterm birth.
Currently, the lack of comprehensive understandings of the uterine contraction initiation mechanism hinders effective early-stage treatment of preterm birth.
 Once delivery starts, it can not be prevented.  Thus, early detection and preemptive treatments are a promising direction for preventing premature babies. Frequently used preterm diagnosis methods include Tocogrametry, Intra-uterine Pressure Catheter, Fetal  Fibronectin, Cervical length measurement etc, but none of theses provides reliable results~\cite{Lucovnik2011}.

The expulsion of a fetus is a direct consequence of strong  periodic uterine contractions, which results from the generation and propagation of action potentials~\cite{Lammers2008}. The corresponding electric signals can be recorded by electrodes placed on the abdomen of pregnant women, using the electrohysterogram (EHG) technique.
%signals typically called electrohysterogram (EHG)
%can be recorded by electrodes placed on the abdomen of pregnant women.
Due to the close relation between uterine contraction and the underlying electrical activities,
EHG provides a new direction for the development of preterm diagnosis method~\cite{Leman1999,Hassan2011}.

%\AP{For the purpose of the present study, we use the publicly available PTEHG database~\cite{Fele2008} that contains}  EHG signals from 300 pregnant women.

Taking advantage of the recent progress in machine learning,
a set of new preterm diagnosis methods have been proposed~\cite{Ren2015, Fergus2016, Acharya2017}.
Overall, the preterm diagnosis can be categorized as a classification problem,
i.e. to decide or to predict a patient (pregnant woman) is at the risk for preterm birth,
based on a set of physical examination data (sample) and the features contained therein.
It is well conceived that
both the abundance of the sample w.r.t. different classes and the quality of the features that distinct different classes
are vital to achieve satisfactory classification results.

In recent years, the TPEHG (Term Preterm EHG) database has been widely used for training and testing variant machine- learning-based preterm diagnosis methods.
Although there are millions of preterm babies world-wide,
the fraction of preterm birth is quite small, compared to the total number of births.
This fact is reflected in the composition of the TPEHG database which is a public available database that contains 300 EHG samples of pregnant women~\cite{Fele2008} .
It is noticeable that only 38 EHG samples are collected from patients whose pregnancies will result in preterm delivery,
while the other 262 EHG samples  are from patients with  normal term delivery.  
Due to the strong difference between the number of preterm and normal delivery samples, applying conventional machine 
%TPEHG database has been widely used for training and testing variant machine-learning-based preterm diagnosis methods.
%Considering the difference in the size of preterm and term samples,
%it is actually a learning task from imbalanced dataset.
learning algorithms with such extremely imbalanced data
will tend to classify the minority samples into the majority class, i.e., there exists a bias towards the majority \cite{Daskalaki2006},
which is likely to result in inaccurate diagnosis result.

% \AP{notoriously} tends to misclassify the minority
% \COM{(bias toward majority, i.e., term )}~,
% thus \AP{making it inappropriate to detect the rare pregnancies that lead to preterm births.}

Learning from imbalanced dataset is very active research topics in the field of machine learning~\cite{He2009,Yan2019}.
The state-of-the-art research methods to deal with imbalanced data mining problem can be categorized into two directions:
1) over-sampling  the minority class or under-sampling  the majority one
in order to compensate the imbalance of samples between classes to be identified;
2) synthesizing artificial samples from minority class.
The former is of limited use  when the size of the dataset is small.%in the case of a limited dataset.
%several constraints  in practical use, especially in the case of limited dataset.
To be specific, under-sampling could significantly reduce the number of samples to be used in training the learning model, potentially leading to under-estimation;
while over-sampling could magnify the feature variations of training samples,
possibly resulting in over-estimation.

On the other side, the synthetic sampling with data generation methods
aims to generate synthetic data that originated from the minority class.
The synthesizing procedure mimics the random distribution of sampling data in the feature space of minority,
so that the generated samples are assumed to be close to the actual distribution of minority in its feature space.
Including these samples as the minority training set eliminates the imbalance in the original dataset,
and removes the classification  bias towards the majority.
The frequently used synthetic algorithms such as SMOTE \cite{Chawla2002} and ADASYN \cite{He2008},
have exhibited certain advantages in real applications of preterm diagnosis~\cite{Fergus2016} and other problems~\cite{Yan2019,Kok2019}.

% For this reason, synthetic data generation are frequently used in machine learning based preterm detection algorithms.
% Classifiers trained with imbalanced dataset typically show bias toward {minority}.
% Synthetic algorithms Although

Since both the abundance of training examples and the quality of features are key factors
to improve the precision of classification~\cite{Andreuperez2015},
it is also important to extract new features from EHG signals,
so that the performance of machine-learning-based preterm prediction algorithms can be improved by combining these new features
\cite{Rabotti2010,Hassan2013,Borowska2018,Shahrdad2018}.
However, notice that %the quantity of the samples and the quality of the features
%interact in a very complex way w.r.t. the classification performance in the case of synthetic sampling.
when new features are adopted in the training process, the effect of imbalance may deteriorate \cite{Blagus2010}.
Also, when more synthetic/artificial data of the minority class are generated, %under the synthetic algorithms,
the representation ability of the features may change. In addition, with the increase of synthetic samples, the noise in the original samples might intensify. Being trained with these dataset, the classifier would overfit~\cite{Hawkins2004}. 
Therefore, adding synthetic samples may affect in a complex way the quality of the features used by the algorithm, and as a result, may alter the classification performances.  Although there are works devoted to optimize synthetic algorithms~\cite{Bunk2012,Nejatian2018}, to the best of our knowledge, few work  concerns the effect of optimal number of synthetic samples to  classifier accuracy.
For this reason,  it is necessary to explore the relation between the amount of synthetic data and the quality of features and
investigate some unified formulation.

% the precision of classifiers might be further improved
% by optimizing the ratio of the number of training examples between classes.
This paper investigates the relation of bias elimination and feature importance reduction when introducing synthetic samples in the training process.
%This paper focuses on the trade-off between the effect of reduced feature importance and bias elimination
%\COM{benefitted from} the introduction of synthetic samples in training process. 
We determine an optimal minority dataset synthesizing strategy by quantifying feature importances and classification precision of synthetic samples.
The rest of the paper is organized as follows:
In Section~\ref{sec:ps}, we introduce some basic factors regarding the problem, and
 analyze the underlying principles of the prevalent synthetic algorithms SMOTE and ADASYN,
from which we demonstrate the importance of finding  an optimal inter-class sample ratio in dealing with learning from imbalanced dataset.  Section~\ref{sec:optimal} further quantifies the effect of synthetic samples  and formulates the problem of determination of optimal sample balance coefficient.
 %the procedure that determine the optimal amount of synthetic samples.
In Section~\ref{sec:IV},
we verify the effectiveness of the proposed method by applying it to EHG based preterm prediction in a numerical way.
Section~\ref{sec:V} concludes the paper.

%
%\blue{(The following two paragraphs are in Section \ref{sec:introduction} of the former manuscript.
%For conciseness, it is possible to delete them.
%However, certain contents of these two paragraphs are adopted in the following statement.)}
%
%(With the increase of synthetic examples, although they originate from  real dataset,
%the noise in the original examples might intensify. Being trained with these dataset,
%the classifier would overfit and give poor performance in real applications~\cite{Hawkins2004}.
%Furthermore, the synthetic examples diminish the bias typically with a price of reducing the classification precision on the majority.
%
%In the case of EHG signal based preterm prediction,
%special attention must be paid when synthesizing data from  EHG signals of the preterm patients.
%On one side,
%enough preterm EHG samples should be synthesized in order to eliminate \AP{the} bias \AP{towards} \COM{normal term (majority)}.
%On the other side,
%classifier precision on majority (normal term) decreases \AP{when artificially increasing the number} preterm examples.
%\AP{The main idea of this work is to suggest that}
%a compromise should be \AP{found} in order to maintain high performance on \AD{both} classes.
%Although \AP{much work has been} devoted to optimize synthetic algorithms~\cite{Bunk2012,Nejatian2018},
%to the \COM{best} of our knowledge, few work concerns the effect of optimal synthetic dataset  in EHG based preterm prediction.)

\section{Problem Statement}\label{sec:ps}

%\blue{(The title of this section deserves further concern. There is no explicit expressions that define/state the problem.)} 

As explained in the introduction, the strong imbalance between pathological and normal outcomes from the available database results in possible inaccuracies in the classification algorithms. 
%Due to the lack of effective treatment of preterm birth,
%early detection and related interventional therapy are highly desirable.
%To this end, diagnosis method should detect patients that are susceptible to preterm
%as precisely and as early as possible.
%At the same time, it should not misclassify the normal term patients to be the preterm suspects.
%Due to the fact that  there are much more normal term cases comparing with the preterm delivery cases, a minority sample is more likely to be mis-classified.
To avoid this, the preferable machine-learning-based algorithms typically introduce a
certain amount of synthetic preterm sample data to mitigate
the bias towards majority (normal delivery). However, the possibility of mis-classifying term samples increase  at the same time. Thus, how many synthetic samples should be added is directly related to the performance of a machine-learning based diagnosis method.

\subsection{Sample balance coefficient and feature score}

As stated in Section \ref{sec:introduction},
the abundance of training samples of all classes is essential to the performance of classification methods.
For instance, the aforementioned TPEHG database contains great more normal term samples than the preterm ones (262:38),
so it is natural to utilize data synthesis techniques to generate samples of the minority class, i.e., the artificial preterm sample.
However, there is a  lack of understanding about
how many minority class samples should be synthesized without changing the number of majority class samples,
and what the after-effect will be w.r.t. the classification performance if we synthesize more than enough samples of minority class.

The enrichment of minority class samples by applying data synthesis techniques will improve the classification performance in certain sense,
however, it is worth noting that
introducing synthetic data might alter the original pattern of sample distribution in the feature space,
depending on the underlying mechanics of synthesis, i.e., the boundary in feature space between different classes might be blurred.
To better quantify the contribution of different features to classification,
we introduce the following feature score $f^i_{s}$ defined as in \cite{Song2017},
\begin{equation} \label{eq:fis}
f^i_{s} =
\frac{ \left( \bar{x}^+_i - \bar{x}_i \right)^2+ \left( \bar{x}^-_i - \bar{x}_i \right)^2}
{\frac{1}{n^+ -1} \sum_{k=1}^{n^+} \left( x^+_{k,i} - \bar{x}^+_i \right)^2
+\frac{1}{n^- -1} \sum_{k=1}^{n^-} \left( x^-_{k,i} - \bar{x}^-_i \right)^2}
\end{equation}
where $x_{k,i}^+$ and $x_{k,i}^-$ denote the measured physical value of feature $i$ from sample $k$ that is in positive (minority or preterm ) class and negative (majority or term ) class, respectively. 
$\bar{x}_i$ is the average value among all samples,
$\bar{x}^+_i$ is the average value of  all $n_+$ positive (minority or preterm) samples, and
$\bar{x}^-_i$ is the average value of $n_-$ negative samples.
According to (\ref{eq:fis}), $f^i_s$ records the discrimination between minority class (+) and majority class (-)
by counting how divergence of   the  samples are in the feature space.
It is worth mentioning that the feature score explicitly relates to the size of testing samples.
We introduce the sample balance coefficient $\alpha$ by,
\begin{equation} \label{eq:alpha}
\alpha = \frac{n_{+}}{n_{-}}
\end{equation}
where $n_{+}$ and $n_{-}$ are the numbers of samples in the minority and majority class
after applying synthetic algorithms, respectively.
Notice that $n_{+}$ include the number of synthesized data.

%\blue{( For consistence, it is possible to use $n^+$ and $n^-$ as in eq. (\ref{eq:fis}). 
%It is also easier for us to use $F_{score}(\alpha)$ in eq. (\ref{eq:def_f_score}). )}

%Most of the time, these data synthesis techniques are used without any distinction, or simply put no discussion to possible differences.

%\blue{(It suddenly came to me that there is no true definition/meaning of $x_i$,
%and $i$ should be labeled as $i=1,\cdots,N$ as in (\ref{eq:def_f_score}).)}

Also, for any specified classification problem,   
different features abstracted from samples jointly contribute to the final classification result.
According to (\ref{eq:alpha}) and (\ref{eq:fis}),
it is reasonable to define the global feature score $F_{score}$ 
as the weighted sum of different feature scores $f^i_{s}$, i.e.,
\begin{equation} \label{eq:def_f_score}
\begin{split}
F_{score}(\alpha) &= \sum_i^{N}w_i\cdot f_{s}^i\\
\sum w_i &=  1
\end{split}
\end{equation}
where the weights $w_i \in [0,1]$  are introduced to represent the importance of feature $i$ to the classification, 
and $N$ is the number of features used in the final classification.
By construction, the definition of (\ref{eq:def_f_score}) links the the number of synthetic samples and the quality of the features.
It provides a unified performance metric which is essential for further investigation.
%Besides, the random forest is used in the rest of the paper to determine each feature's weight $w_i$
%since it is capable to return certain measures of feature importance \cite{Diazuriarte2006}. 
%\blue{(The following sentence can be stated in Section IV as the experiment procedure.)}
To proceed, all features are initially used to build a forest, from which we obtain a value of classification accuracy~\cite{Diazuriarte2006}.  
The reduced classification accuracy by randomly permuting a node in the tree gives a reliable measure of the feature's importance\cite{Breau2003}.

\subsection{Size of the synthesized data and distinguishability of the features}

After introducing the definitions of sample balance coefficient $\alpha$ and feature score $f^i_s$ / $F_{score}(\alpha)$,
it is convenient  to investigate the attributes of the conventional data synthesis algorithms such as SMOTE or ADASYN,
and consider their feasibilities in the application of preterm diagnosis using the TPEHG database.

%While the Minority data-set synthesis  algorithms, like SMOTE or ADASYN, tend to mimic the natural distribution of sample in its feature space.
%To this end, they are widely used in tackling learning tasks from imbalanced dataset~\cite{8702462,8642396}.

\begin{figure}[h]
	\centering
	\includegraphics[width=0.4\textwidth]{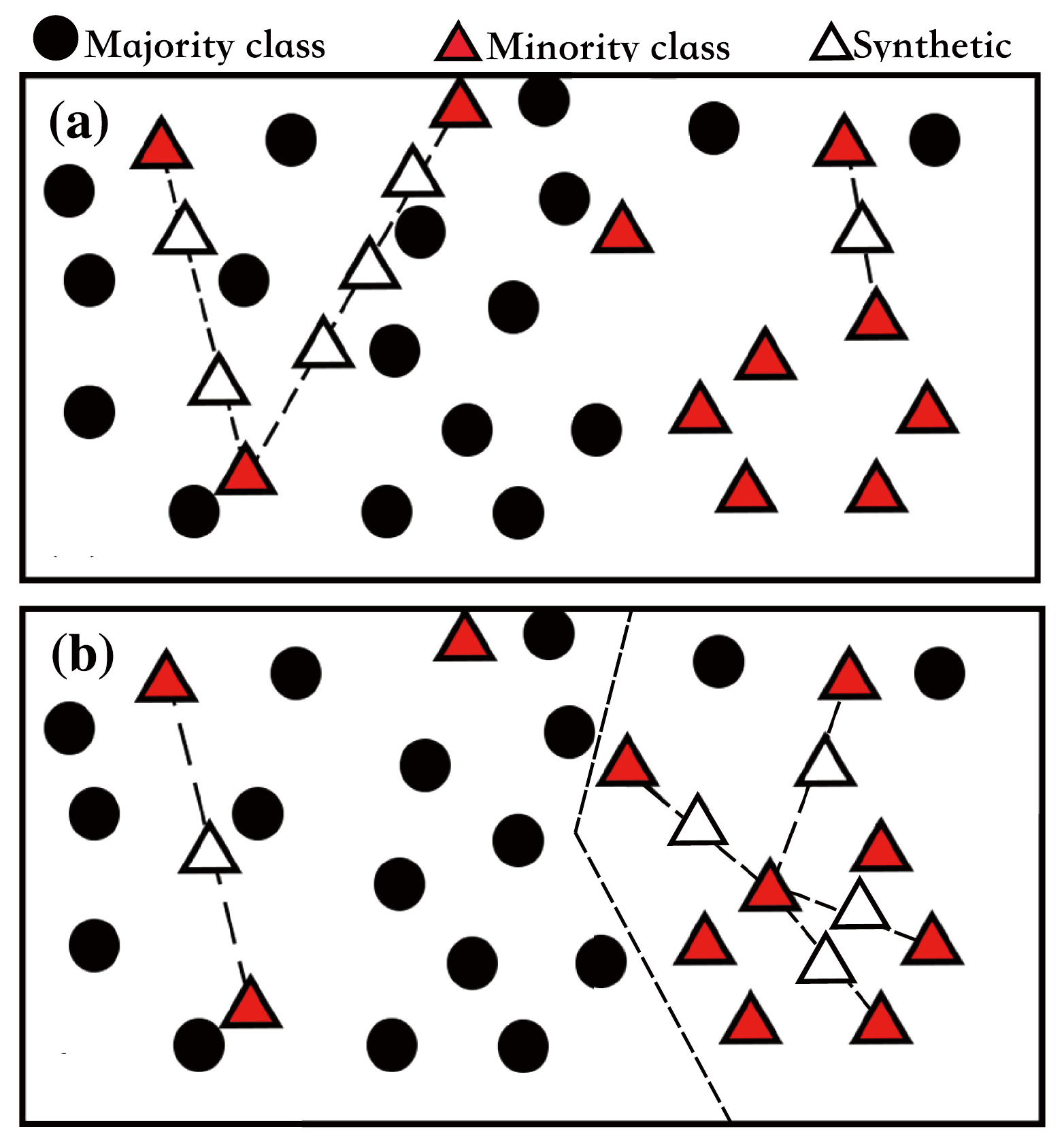}
	\caption{Schematic illustration of sample synthetic algorithms.   (a) ADASYN tends to synthesize more  artificial samples with the original minorities surrounded with more majority samples. It might cripple the original separability  of samples.  (b) SMOTE algorithm synthesize data point with a randomly chosen minority sample in its closest neighbors. It is easier to keep the original sample distribution in feature space. }
\label{fig:synthetic_method}
\end{figure}

Although data synthesis algorithms tend to mimic the natural distribution of sample in its feature space,
the nature of random synthesis of minority samples inevitably has certain effects on features' ability to discriminate different classes.
For instance, ADASYN tries to synthesize more data from  minority samples surrounded by more majority~\cite{He2008}.
As shown in Fig.\ref{fig:synthetic_method} (a),
the synthetic samples are more likely to appear on the left due to more majority data samples around each minority ones.
Although they intend to ease classification by emphasizing on samples that are hard to learn from,
at the same time, they would potentially make originally separable datasets non-discriminated in the feature space.

Contrary to ADASYN, SMOTE does not take into account the  surrounding of the minority samples.
For any minority sample $s_i$, it randomly selects another minority  $s_k$ among its $k^{\text{th}}$ nearest neighbors
and synthesizes an artificial samples $s_s= s_i +\lambda(s_k-s_i)$ with a random number $\lambda \in[0,1]$.
As a result, the synthesized samples will concentrate in the region containing more minority samples (see Fig.\ref{fig:synthetic_method}(b)),
which implies that it maintains the original distribution pattern, without diminishing features' contribution to classification.

To see the effect of synthetic sampling  on features' contribution to classification,
we apply the aforementioned two synthetic algorithms to TPEHG database with frequently used features:
1) the root mean square value of the signal ($rms$);
2) the median ($F_{med}$) and peak ($F_{peak}$) frequency of the power spectrum;
3) the sample entropy of the signal ($E_{samp}$) extracted from EHG signals. % recorded in TPEHG database\cite{Fele2008}.
We use hose with the gestation age less than 37 weeks as the preterm (minority class) samples.

% the recorded gestation age with the criteria less than 37 weeks
% to separate preterm samples,
% which gives us a minority (Preterm) class  of 38 samples while 282 samples belonging to the majority (Term) class.

\begin{figure}[h]
\begin{center}
\includegraphics[width=0.5\textwidth]{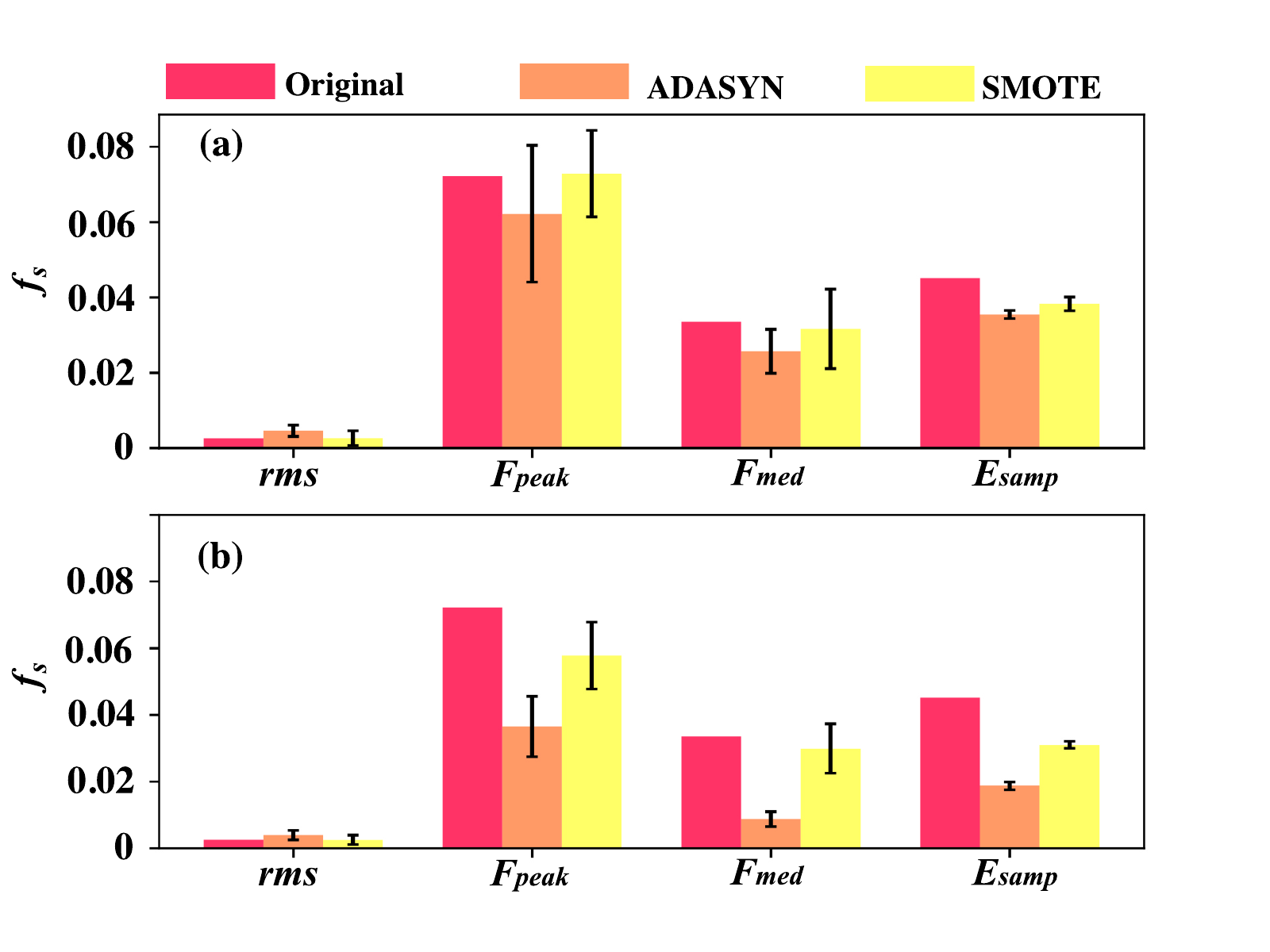}
\caption{Effect of synthetic samples on features' contribution to class separation measured using feature score $f_s$.  Both synthetic methods weaken features importance with the increase of synthetic samples. (a) Sample balance coefficient  $\alpha=0.3$, (b) $\alpha =0.5$  }
\label{fig:synthetic_effect}
\end{center}
\end{figure}

Fig.~\ref{fig:synthetic_effect} shows the variation of features' contribution to classification represented by its $f^i_s$ after synthetic sampling.
It is evident that the peak frequency gives the highest feature score $f^i_s$ among these four features.
Its effectiveness for classification has been confirmed by other authors~\cite{Lucovnik2011,MANER2003, Fergus2013}. 
Also, it is quite astonished to see that both techniques tend to deteriorate features' ability in separating samples. 
SMOTE shows more superiority, i.e., the feature scores are higher after applying SMOTE than those after applying ADASYN. 
This is consistent with the previous analysis of data synthesis mechanics of SMOTE and ADASYN.

% However, its effectiveness decreases with the introducing of synthetic samples. Same trend appears for other features. 

%we introduce sample balance coefficient $\alpha$ defined as
%\begin{equation}
%\alpha = \frac{N_{min}}{N_{maj}}
%\end{equation}
%where $N_{min}$ and $N_{maj}$ are the total number of samples in the minority (Preterm) and majority (Term) class
%\COM{after applying synthetic algorithms}, respectively. 

It is also worth noting that features' classification ability is sensitive to the number of synthetic samples introduced. 
This can be seen from Fig.~\ref{fig:synthetic_effect}, which shows the 
featured scores $f_s^i$ after adding synthetic data with 
different sample balance coefficients $\alpha$ (see Eq.~\eqref{eq:alpha})
$\alpha = 0.3$, panel (a) and $\alpha = 0.5$, panel (b).
As shown by the figure, adding synthetic samples degrades the capability
of the algorithm to distinguish the various features.
%To see this effect, recall the sample balance coefficient $\alpha$ (\ref{eq:alpha}).
%We see that Fig. \ref{fig:synthetic_effect} gives feature scores $f^i_s$ after applying the synthetic algorithms
%with different sample balance coefficients,  $\alpha=0.3$ in panel (a), and $\alpha=0.5$ in panel (b). 
%It is quite evident that the more the synthetic samples are added, the less effective the features are. 
However, synthetic samples are required to eliminate classification bias against the minority. 
As a result, a trade-off should be found between the number of synthetic samples and features' quality, 
in order to optimize the final performance of classifiers trained with these data.

\section{Determination of optimal sample balance coefficient} \label{sec:optimal}

As discussed in Section \ref{sec:ps}, to accomplish a machine-learning-based preterm diagnosis,
synthetic sampling with data generation of the minority class is inevitable,
and the balance between synthesized data and the feature quality must be considered.
Intuitively, increasing the number of minority samples by generating synthesized data 
should increase the prediction precision on  the minority class and reduce the bias towards the majority. 
On the other hand,  the prediction precision on majority class may fall if there exists too many minority samples.
Ideally, we would expect a no-bias learning system when the sample balance coefficient $\alpha=1.0$.
In real applications, however, due to the imbalance of the available original samples between classes,
the optimum may differ from $\alpha = 1$.
To this end, we introduce two functions $C_+$ and $C_-$ describing the putative biases 
induced by the sampling on the minority and the majority class, respectively,
\begin{equation} \label{eq:act}
\begin{split}
C_+(\alpha) = &\frac{1}{1+\exp(-k(\alpha-1+\overline{\alpha}_0))}\\
C_{-}(\alpha) =& \frac{1}{1+\exp(k(\alpha-1+\overline{\alpha}_0))}\\
\overline{\alpha}_0 =& \alpha_0(1-\alpha_0)
\end{split}
\end{equation}
In the present work, we chose the parameter $k = 4$ to describe
how the prediction is affected by the balance coefficient $\alpha$.
In Eq.~\eqref{eq:act}, $\alpha_0$ is the original sample balance
coefficient before generating any synthetic data.

Eq (\ref{eq:act}) implies that 
when $\alpha$ is smaller than $1$, the majority is well described, and $C_-$ is close to $1$. 
On the other hand, when $\alpha$ is large, the minority is accurately described ($C_+$ is close to $1$), 
but the majority will be affected ($C_-$ is reduced).

%$\alpha$ has a finite value, towards the accuracy in describing the
%  This inspires us to introduce the activation $C_+$ and inactivation function $C_-$ of a learning system defined as~\cite{Rihana2009}
% , the} function $C_+(\alpha)$ describes how the sample balance coefficient $\alpha$ affects the prediction precision on majority sample; and $C_-(\alpha)$ describes the effect on minority sample prediction.

For most of the classification purposes, 
a high precision on different classes is demanded (no bias towards any class).  
To this end, in the case of learning problems from imbalanced data samples, 
it is required that minority training examples are re-sampled to match the number of majority, i.e., $\alpha=1$, 
in order to remove potential bias towards majority. 
However, recall the synthesis procedure, 
additional training examples are generated from the original minority data sets 
and can be seen as a kind of ``copy'' of the original data. 
As a consequence, adding too many synthetic data improves only the precision on the minority class, while degrading the accuracy on the majority classification.

%For an ideal learning system\AP{, an implicit assumption is that $\alpha = 1$
%provides the optimal choice.} In this case,
%where the activation and inactivation functions show no-bias toward any class, $C_+=C_-=0.5$.
%However, when the training samples are imbalanced ($\alpha<1$), classifiers show poor performance on minority sample prediction (~low value of $C_-$~), while high precision on majority (~high value of $C_+$~).

To account for these effects, 
we introduce the constant $\overline{\alpha}_0$ in the definitions of $C_{+}$ and $C_-$ in Eq.\eqref{eq:act}. 
This equation expresses that when there are enough original minority samples ($\alpha_0\rightarrow 1$),  
constant $\overline{\alpha}_0$ goes to 0. 
Thus there is no need to synthesize training examples. 
When learning tasks from imbalanced dataset are encountered ($\alpha_0 < 1$),  
the optimal balance coefficient $\alpha$  deviates from the ideal value.

To take into account the bias described above, we simply multiply the effective feature score, $F_{score}$, 
as defined in Eq.~\eqref{eq:def_f_score}, by $C_{+} \times C_{-}$, to come up with an effective score, $F_{score}^e$.
Combining the requirement on high prediction performance on preterm as well as term, 
%we introduce the effective feature score $F_{score}^e$, through which
we determine the optimal sample balance coefficient $\alpha^*$ as follows,
\begin{equation} \label{eq:op-alpha}
\begin{split}
\alpha^*=&\text{argmax}_{\alpha\in[0, \inf]}F_{score}^e(\alpha)\\
F_{score}^e(\alpha)=& F_{score}(\alpha)\cdot C_+(\alpha)\cdot C_-(\alpha)
\end{split}
\end{equation}

\section{Experiment verification}\label{sec:IV}

\begin{figure}[htbp]
\begin{center}
\includegraphics[width=0.5\textwidth]{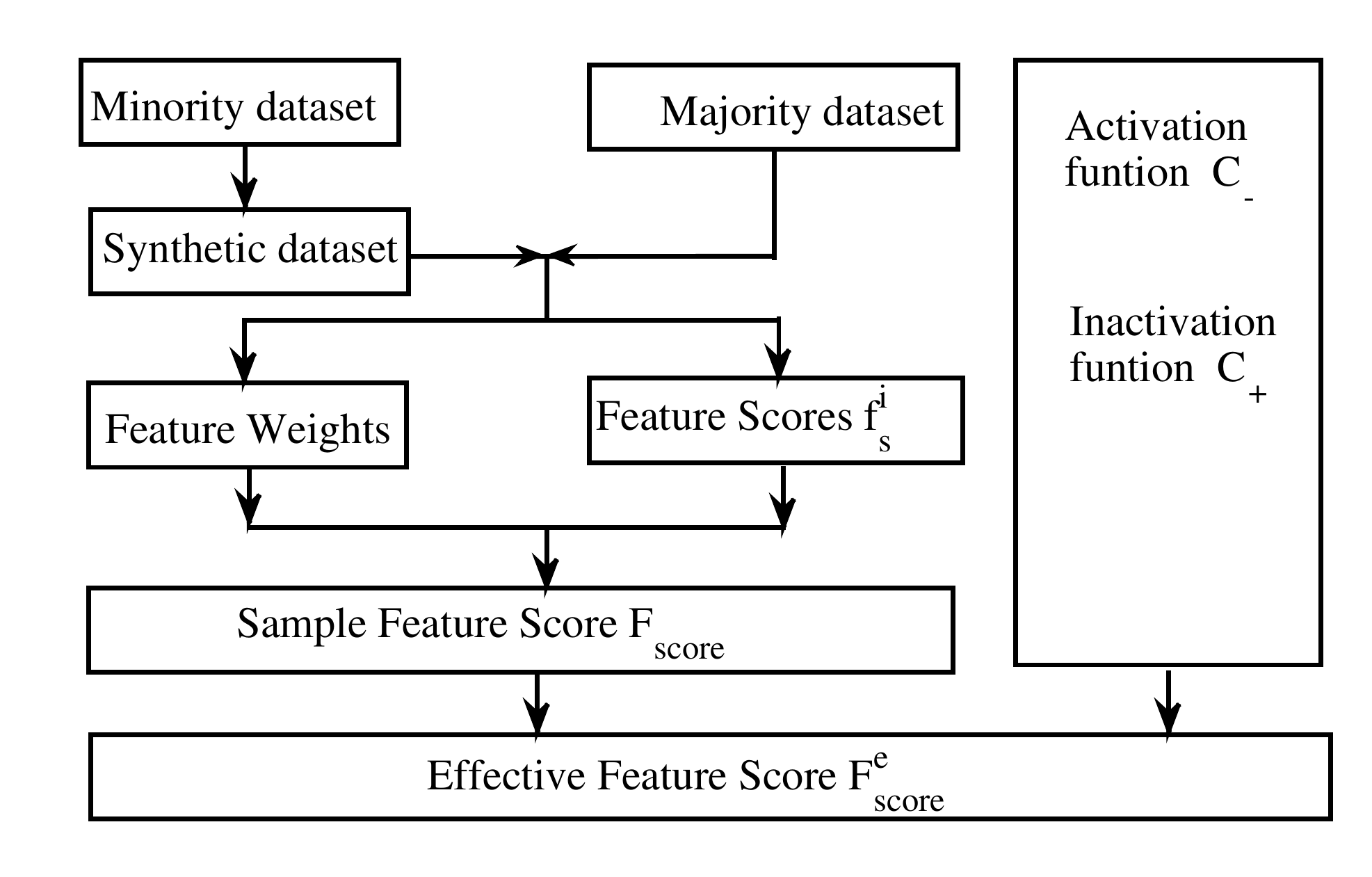}
\caption{Flowchart for determining effective feature scores.  After having synthesized artificial samples, calculate the weight $w_i$ and score value $f^i_{s}$ of each feature. Based on these two values, the weighted  feature score $F_{score}$ is defined.  Combining the activation/inactivation functions with $F_{score}$ gives effective feature score $F^e_{score}$.  }
\label{fig:optimal_balance_selection}
\end{center}
\end{figure}

The above analysis suggests a way to improve preterm diagnosis precision 
by determining the optimal sample balance coefficient without sacrificing the precision on term prediction.
In this section, we provide experimental results to verify the effectiveness of the proposed method, 
particularly, we propose the numerical way to the determination of $\alpha^*$.
The general procedure regarding the experiment is summarized in Fig. \ref{fig:optimal_balance_selection},

\subsection{Material and Terminologies}

Electrohysterogram (EHG) data obtained from the Physionet database (TPEHG) are used. 
Root mean square ($rms$), peak ($F_{peak}$) and median ($F_{med}$) frequencies and sample entropy ($E_{samp}$) are extracted from the recorded EHG signals.  Based on the recorded gestation age ($T_{gest}$), 300 samples are spilt into two groups with the criteria $T_{gest}<37$ weeks, which gives a minority class ( preterm ) of 38 samples and a majority group (term) of 262 samples.

Recalling the purpose of predicting gestation status,  
frequently used simple but powerful classifiers, 
like Support Vector Classifier (SVC),  Linear Discriminant Classifier (LDC), Logistic Regression Classifier (LRC), 
Decision Tree Classifier (DTC), Gradient Boosting Classifier (GBC) etc., are used to verify the proposed method.  
Following \cite{Fergus2013}, 
Holdout Cross-Validation with 80\% of the whole dataset is designated for training the classifiers and the rest 20\% for testing, 
from which we calculate sensitivity (True Positive (preterm) Rate, TPR) and specificity (True Negative (term) Rate, TNR).  
Notice that these two quantities alone can not well represent our requirement on high performance 
of both positive and negative prediction. 
For this purpose, we introduce $G_{mean}$ and Overall Accuracy ($OA$) as the performance metrics as follows,
\begin{equation}\label{eq:gmean}
\begin{split}
G_{mean}&=\sqrt{\text{TPR}\times\text{TNR}}\\
OA &= \frac{\text{TP+TN}}{\text{TP+FP +TN +FN}}
\end{split}
\end{equation}
where \text{TP} and TN are correctly predicted positive and negative test samples, 
\text{FP} and FN are incorrect predictions, respectively. Apparently, 
classifiers provide the most accurate prediction on both preterm and term classes 
are typically with the highest value of $G_{mean}$ and overall accuracy $OA$. 
Beside these quantities, Area Under (Receiver Operator) Curve (AUC) \cite{Fawcett2006} is also used to verify the proposed method.

\subsection{Optimal synthetic preterm samples}

Based on the intuitive analysis given in the previous sections,  SMOTE has stronger capability than ADASYN in keeping features importance in classification. We first
apply it to generate synthesized samples of minority class to match a given  sample balance coefficient $\alpha$.  

\begin{figure}[h]
\begin{center}
\includegraphics[width=0.5\textwidth]{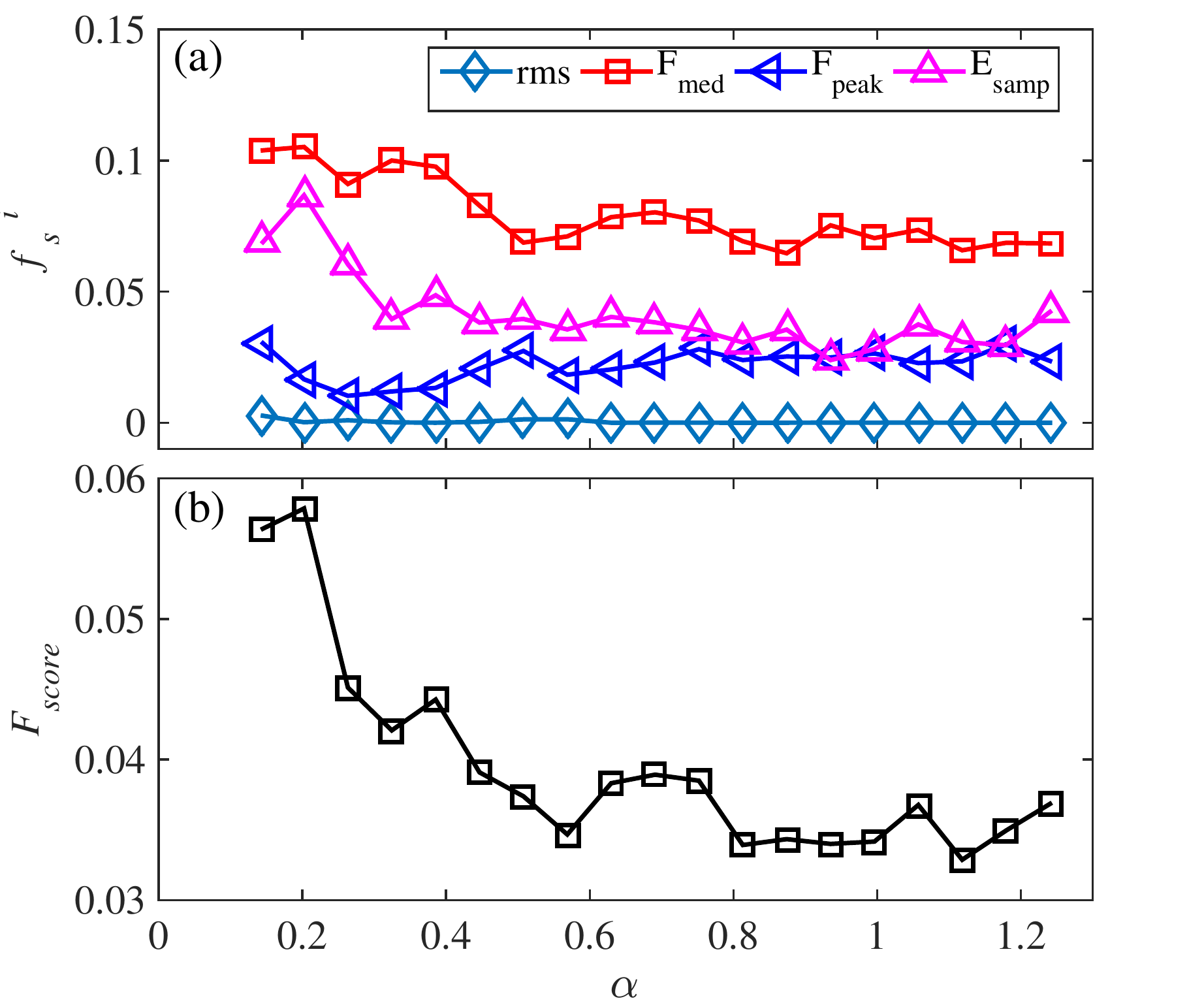}
\caption{Measured feature scores $f^i_{s}$ (a)  and $F_{score}$ (b) at different sample balance coefficient $\alpha$. The importances of  each of the  frequently used  features $f_{s}^i$ show variation after applying synthetic algorithm SMOTE. However,  the global feature score $F_{score}$ obtained from the importance $w_i$ and feature score $f_{s}^i$ shows continuous decrease with the increase of $\alpha$.  }
\label{fig:feature_score}
\end{center}
\end{figure}

Fig. \ref{fig:feature_score}(a) shows the measured feature score $f_s^i$ with different $\alpha$.  
Obviously, it is hard to determine how many synthetic samples should be generated, 
as features have their measured $f_{s}^i$ values well separated and vary with $\alpha$.

\begin{figure}[htbp]
\begin{center}
\includegraphics[width=0.5\textwidth]{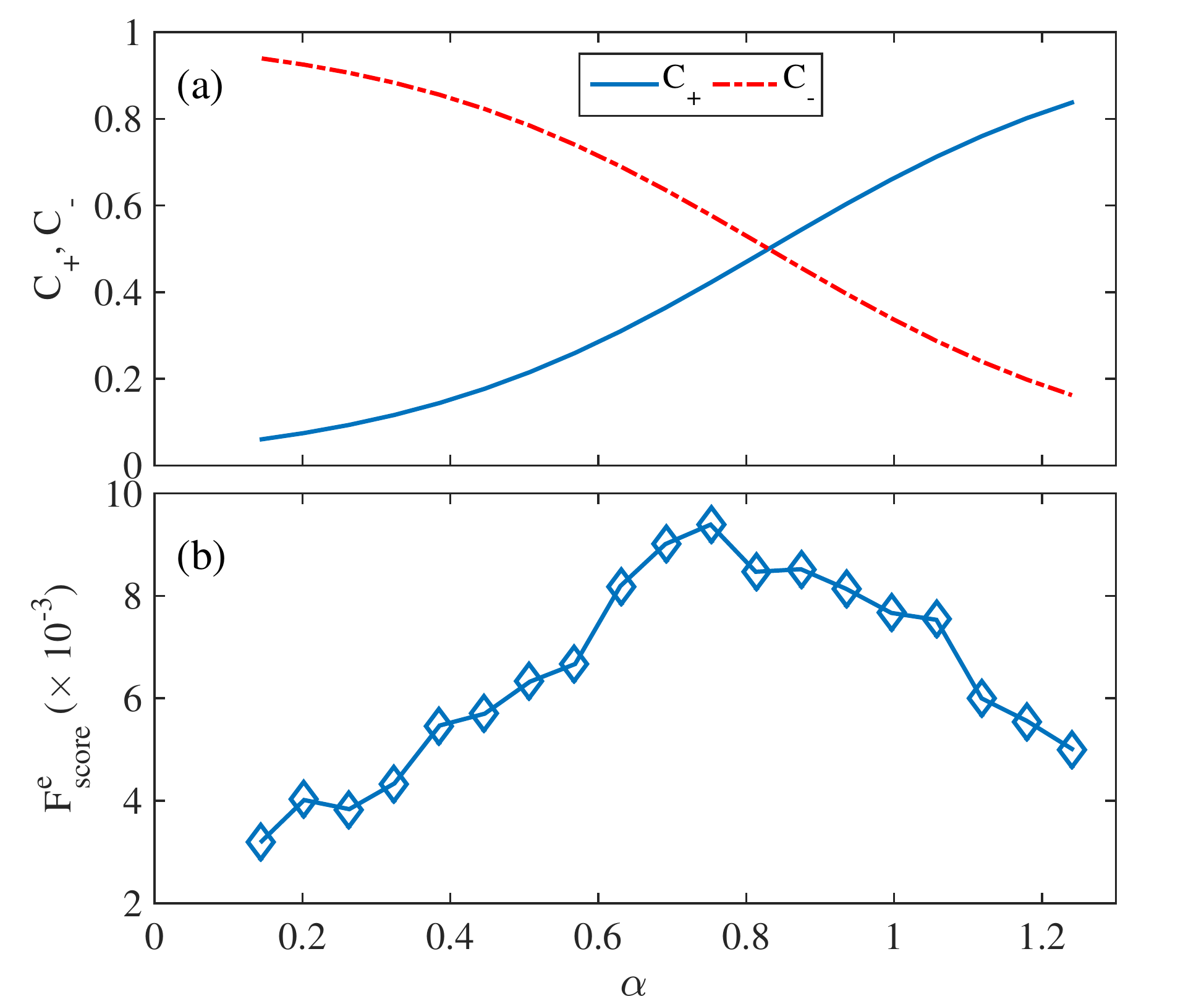}
\caption{Variation of the  effective feature score $F_{score}^e$ ( panel b) calculated from the activation and inactivation function (panel a)  at different sample balance coefficient $\alpha$. $F_{score}^e$ shows a peak at $\alpha\approx 0.7$, which determines the optimal sample balance coefficient $\alpha^*$. }
\label{fig:optimal_alpha}
\end{center}
\end{figure}

After having synthesized enough artificial samples, we  use the features to build a forest, from which we obtain a value of classification accuracy.  
The reduced classification accuracy by randomly permuting a node in the tree gives a reliable measure of the feature's importance (weight)
\cite{Breau2003}. Calculating features' importance (weight) at each $\alpha$ 
allows us to examine the combined effect of synthetic samples on features' importance $F_{score}$. 
As shown in Fig.~\ref{fig:feature_score} (b), $F_{score}$ decreases with the increase of synthetic samples, 
illustrating the drawbacks of the synthetic sampling.  
This also implies the importance of the determination of the optimal sample balance coefficient $\alpha$.

Combining the activation and inactivation functions previously introduced ( Fig.~\ref{fig:optimal_alpha}(a) ), 
the effective feature score $F^e_{score}$ shows a trend that helps us  to easily determine out the optimal sample balance coefficient $\alpha^*$. 
As shown in Fig.~\ref{fig:optimal_alpha}(b), 
$F_{score}^e$ firstly increases with the increase of $\alpha$, 
manifesting the effect of synthetic samples for eliminating bias towards the majority (term) samples. 
Due to the weakening of features' scores and the bias towards term at large $\alpha$, 
when $F_{score}^e$ reaches its peak at $\alpha\approx 0.7$, it starts to decrease.
The position where the peak $F_{score}^e$  locates provides the optimal $\alpha^*$.  
Fig.~\ref{fig:optimal_alpha} demonstrates that with the optimal $\alpha^*$, 
the capability of the various features to distinguish between
different classes has not been lost, while reducing the bias towards the 
majority.
%Apparently, with the optimal $\alpha^*$, 
%features don't sacrifice their capabilities of distinguishing samples in order to remove bias toward majority.

\subsection{Validation}

\begin{figure}[h]
\begin{center}
\includegraphics[width=0.5\textwidth]{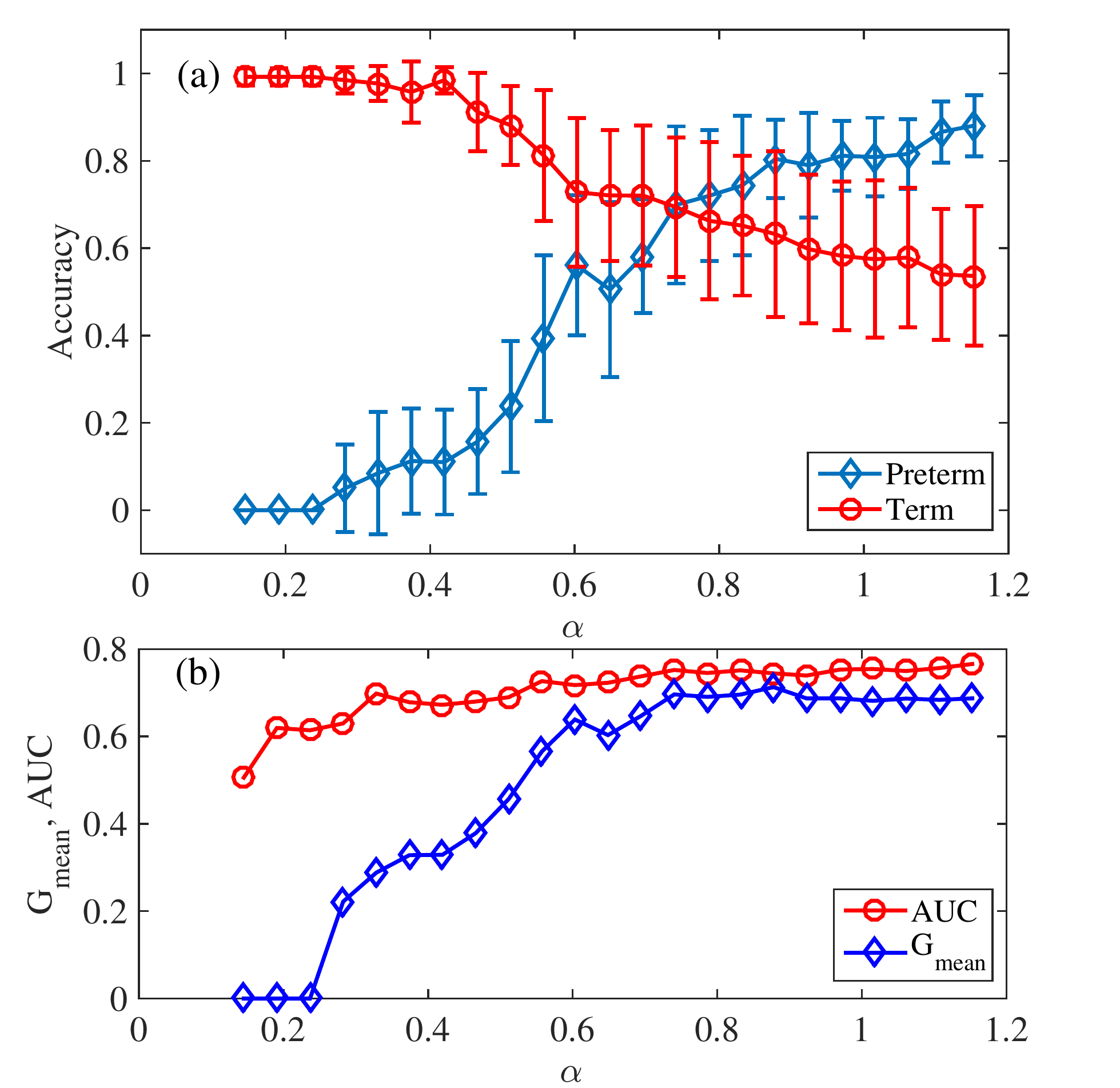}
\caption{Prediction performance of SVC classifier at different sample balance coefficient $\alpha$. The SVC is trained with 80\% of 262 term  and $262\cdot\alpha$  preterm samples. Prediction results are obtained with the rest of 20\% samples. The solid curves are averaged true positive ( preterm prediction precision) and true negative rate ( term prediction).    The inner panel shows the variation of $G_{mean}$ and AUC (area under curve) with respect to sample balance $\alpha$. }
\label{fig:svc_performance}
\end{center}
\end{figure}

To verify the obtained optimal sample balance coefficient $\alpha^*$,  
same features extracted from  80\% of samples in TPEHG database are used to train a SVC classifier, 
the other 20\% are then used for verification. %\blue{(This sentence is quite repeated compared with subsection IV.A.)}
As training and testing samples are randomly selected,  
the calculated quantities representing classifiers' performance vary from time to time.  
For this reason, we repeat the training-testing process for 100 times at each $\alpha$.  
Fig.~\ref{fig:svc_performance} shows the variation of the calculated system performance 
with the increase number of synthetic samples added to the dataset, i.e., the increase of $\alpha$.  
As expected, prediction precision on minority increases, while that of majority decreases. 
It is worth noting that the two curves intersect at the point $\alpha\approx 0.7$, 
which is the optimal sample balance coefficient $\alpha^*$ determined previously.  
At this point the trained classifier eliminates most of the bias toward the majority (term)
 and increases the precision on minority (preterm) prediction 
without sacrificing too much on the precision of term prediction.  
This is confirmed by the accompanied variation of $G_{mean}$ and AUC, 
see Fig.~\ref{fig:svc_performance}(b), where both of these two quantities reach their apex at $\alpha\approx \alpha^*$.

\begin{figure}[h]
\begin{center}
\includegraphics[width=0.5\textwidth]{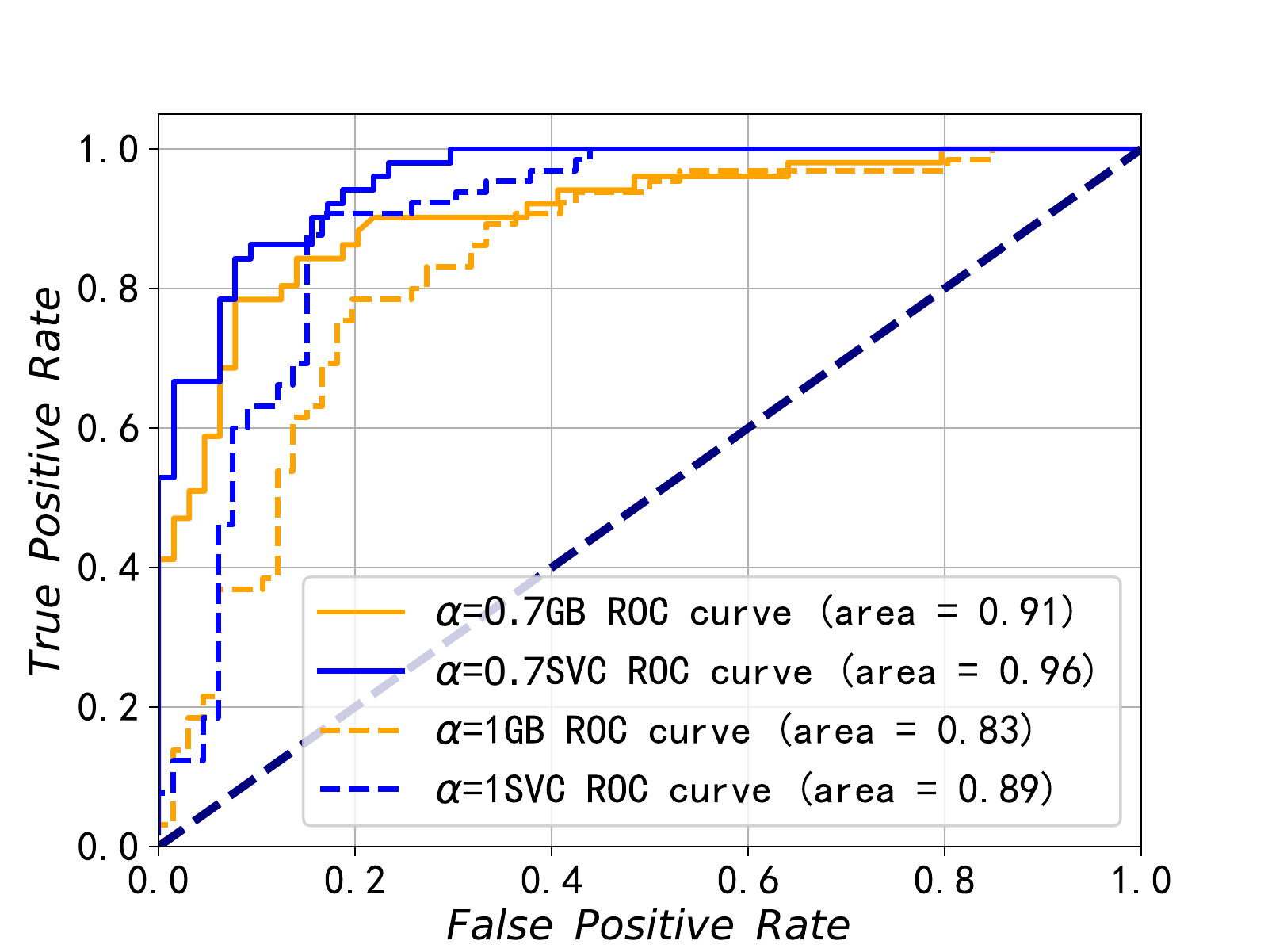}
\caption{Receiver Operator Curve (ROC) for SVC ( blue ) and GBC (yellow) classifiers under different sample balance coefficient $\alpha$. In the case of optimal sample balance coefficient $\alpha^*$(=0.7, determined using the proposed method, solid line), classifiers show better performance.}
\label{fig:roc}
\end{center}
\end{figure}

The receiver operator characteristic curve (ROC)  and the associated AUC values shown in Fig.~\ref{fig:roc} 
indicate the cut-off values for the true positive and false positive rates at different sample balance level (different $\alpha$). 
Apparently, in the case of  optimal sample balance coefficient $\alpha_0$, 
the SVC classifier shows better performance. 
Comparing to the case of ideal balance ($\alpha=1.0$), 
training with optimal amount of synthetic samples leads to a big improvement in terms of AUC values.

\begin{table}[h]
\caption{Comparison of classifier performance under optimal sampling}
\begin{tabular}{|c|c|c|c|c|c|c|}
\hline
\multirow{2}{*}{Classifier} &
\multicolumn{3}{c|}{$\alpha=1.0$} &
\multicolumn{3}{c|}{$\alpha=\alpha^*$} \\
\cline{2-7}
  & OA& $G_{mean}$ & AUC& OA& $G_{mean}$ & AUC\\
\hline
LDC & 0.62 & 0.62 & 0.66& {\bf 0.68} & {\bf 0.65} & {\bf 0.71}  \\
%\hline
%SVM & 0.78 & 0.78 & 0.88 & {\bf 0.82} & {\bf0.83} & {\bf 0.96}  \\
\hline
DTC & 0.75 &0.74 & 0.83 & {\bf0.82} &{\bf0.82} &{\bf0.86}\\
\hline
GBC & 0.78 & 0.78 & 0.83 &{\bf0.85}&{\bf0.84}&{\bf0.91}\\
\hline
\end{tabular}
\label{t:performance}
\end{table}

The advantages of determining the optimal sample balance coefficient shown in SVC  hold for other  different classifiers. 
Table~\ref{t:performance} gives a comparison of frequently used parameters for evaluating classifiers' performance. 
It can be observed that with the previously determined optimal sample balance coefficient $\alpha^*$, 
all the classifiers show a great improvement in performance, especially for SVC based classifiers.

\begin{figure}[t]
\begin{center}
\includegraphics[width=0.5\textwidth]{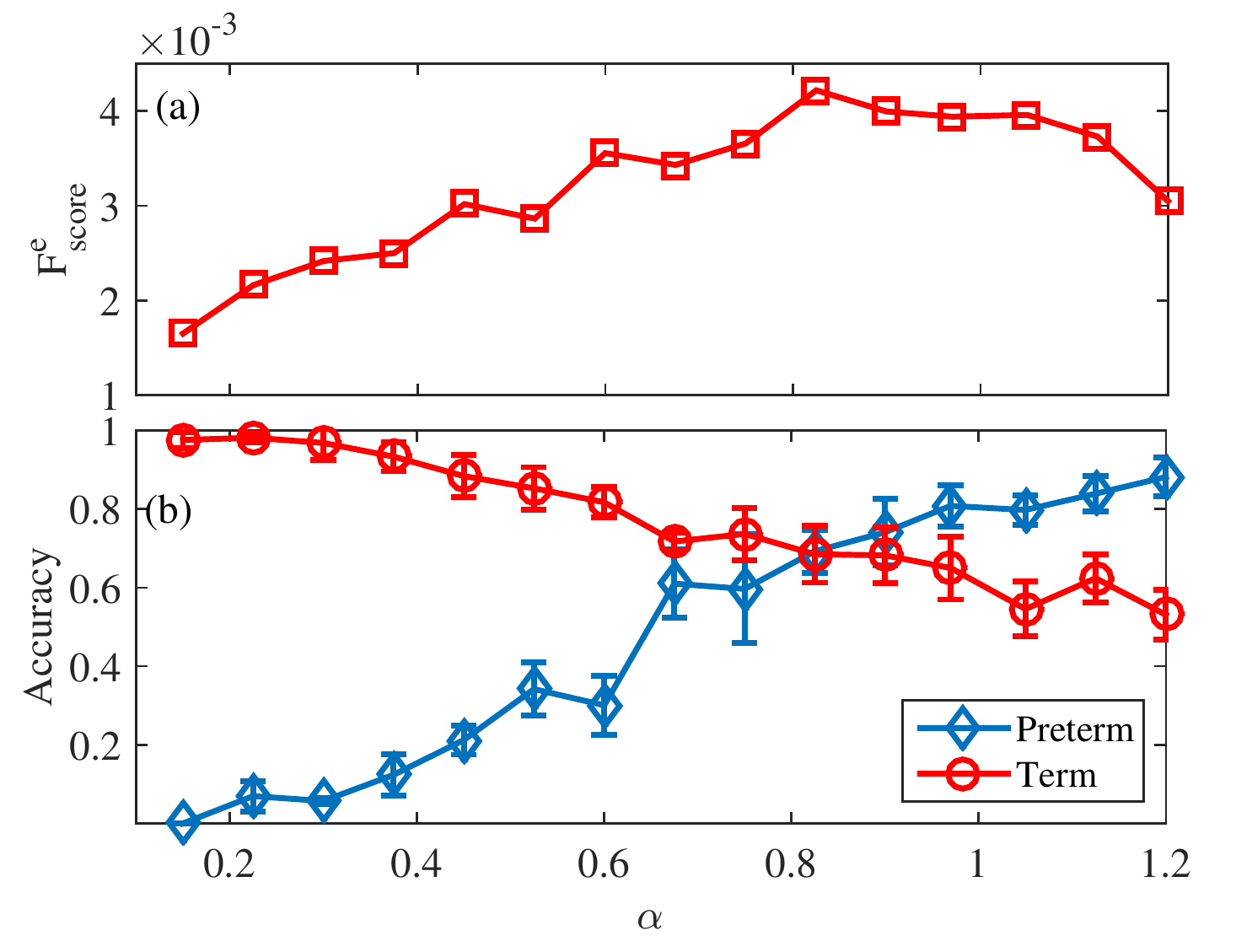}
\caption{(a) Variation of effective feature score $F^e_{score}$ with respect to sample balance coefficient $\alpha$. Artificial samples are synthesized using ADASYN to give sample balance coefficient $\alpha$.  $F^e_{score}$ shows a peak at $\alpha^*\approx 0.8$. (b) Prediction precision  of SVC classifiers on term and preterm after been trained with different amount of synthetic samples using ADASYN method. }
\label{fig:adasyn}
\end{center}
\end{figure}

The effect of optimal sample balance coefficient also works with ADASYN.  Although this method has less power in keeping features' ability in classification,  combining the proposed the activation and inactivation functions do give us an easy-to-identify optimal sample balance coefficient $\alpha_* = 0.8$. As shown in Fig.~\ref{fig:adasyn}, the term and preterm prediction accuracy of a SVC classifier  trained from these dataset  gives the optimal performance at this $\alpha_*$.    However, as effective feature score $F_{score}^e$ 
obtained from ADASYN is less than that from SMOTE as indicated in Fig.\ref{fig:optimal_alpha}, 
it can be expected that the corresponding classifier after training through ADASYN method is underperformed than that through SMOTE method.

%Applying the proposed method to data samples synthesized using ADASYN gives $\alpha^* = 0.8$.  
%Fig.~\ref{fig:adasyn} shows the variation of term/preterm prediction accuracy of SVC classifier 
%after \blue{synthesizing samples under different sample balance coefficient $\alpha$}.  
%Due to the fact that synthetic samples from ADASYN has lower ability in  keeping features' ability of class separation, 
%the intersection point appeared at $\alpha\approx \alpha^*$ has lower accuracy values.

%\blue{(There are so many experiments are conducted in this part!
%However, I think the crucial point is to emphasize certain things about ``numerical determination of optimal sample balance coefficients''.)}
%

\section{Conclusion \& Discussion}\label{sec:V}

Machine-learning based automatic disease diagnosing system provides a prospective direction for modern healthcare. In these applications,  the availability of healthcare data and the effectiveness of  features  extracted from these samples play a crucial role. However, healthcare data are typically imbalanced, with most of samples being healthy (majority), and a few being disease (minority).  Training with imbalanced dataset, classifiers typically introduce  biases towards the majority, making the automatic diagnosing system less useful. To solve the problem, data synthesizing algorithms are used to generate artificial samples from the minority. However, this is typically accompanied by reduced ability of features in class separation.  In this paper,  we propose a method for determining the optimal number of artificial samples should be synthesized. To proceed, we measure features' contributions and their weights in  class separation in the case of introducing different amount of synthetic samples.  Combining with the activation and inactivation functions introduced to describe the effect of sample abundance on  classification precision, we obtain the optimal sample balance coefficient that compromises the effect of  synthetic samples on eliminating bias and  the side-effect of weakening feature importance.  We apply the proposed method to predict preterm behavior using features extracted from public available database TPEHG.   After applying synthetic algorithms, system performances are compared under different scenarios and the results highlight the importance of optimal sample balance coefficient proposed in the work.
%between that  trained with frequently adopted ideally and that of optimally balanced dataset after applying synthetic algorithms.  Highest values of parameters evaluating system performance  obtained at the optimal sample balance coefficient verify the proposed method.

One might argue that it is more critical that an automatic  diagnosis system  mis-identifies a real preterm patient than that mis-identifies a term patient, considering the consequences of serious complexities the preterm babies would have. As such, increasing synthetic samples should be of greater interest, which is the case shown in Fig.\ref{fig:svc_performance} and \ref{fig:adasyn}.  However, special attention should be paid before drawing this conclusion.  Since there is no practical test for any EHG based preterm diagnosis system,  its performance is typically verified using data samples randomly chosen from the total sample set.  The dataset  used to check the performance on preterm prediction are synthesized from the same minority class as those used for the training purpose. With the increase number of  synthetic samples included, the validation samples are getting closer and closer to the training samples, which lead to  unreal high values of preterm prediction accuracy, especially in real applications\cite{Vandewiele2019}.  In the proposed method,  by introducing the activation/inactivation functions that account for the original size of minority samples, we suppress this side-effect.
It is believed that validation results should be close to really applications.

\ifCLASSOPTIONcaptionsoff
  \newpage
\fi



\bibliographystyle{IEEEtran}
\bibliography{ref}

\end{document}